\documentclass[twoside]{article}

\usepackage[accepted]{aistats2025}
\usepackage{tikz}
\usetikzlibrary{shapes,positioning,fit}
\usetikzlibrary {shapes.geometric}
\definecolor{airforceblue}{rgb}{0.36,0.54,0.66}
\usepackage[utf8]{inputenc} 
\usepackage[T1]{fontenc}    
\usepackage[hidelinks]{hyperref}   
\usepackage{xcolor}
\hypersetup{
    colorlinks,
    linkcolor={red!50!black},
    citecolor={blue!50!black},
    urlcolor={blue!80!black}
}
\usepackage{url}            
\usepackage{booktabs}    
\usepackage{dsfont}

\usepackage{nicefrac}       
\usepackage{microtype}      
\usepackage{xcolor}         
\usepackage{tcolorbox}
\usepackage{amsfonts,amsmath,amssymb,amsthm}
\usepackage{multicol}
\usepackage{multirow}
\usepackage{algorithmic}
\usepackage{algorithm}
\usepackage{caption}
\usepackage{tabularx}
\usepackage{wrapfig}
\usepackage{xcolor}
\usepackage{todonotes}
\newtheorem{proposition}{Proposition}

\newtheorem{corollary}{Corollary}
\newtheorem{remark}{Remark}

\usepackage{tikz}
\definecolor{pastelPink}{RGB}{250, 188, 63}
\definecolor{pastelBlue}{RGB}{232, 92, 13}
\definecolor{pastelGreen}{RGB}{199, 37, 62}
\definecolor{pastelYellow}{RGB}{130, 17, 49}

\usetikzlibrary{fadings, decorations.pathmorphing, decorations.markings, shadings}

\tikzset{
  brownian/.style={
    decoration={
      markings,
      mark=at position 0.5 with {\arrow{>}}
    },
    decorate,
    decoration={random steps,segment length=0.75pt,amplitude=1.5pt},
    thin
  }
}

\tikzfading[name=fade out, inner color=transparent!0, outer color=transparent!100]
\usepackage[round]{natbib}

\begin{document}

\twocolumn[

\aistatstitle{Parabolic Continual Learning}

\aistatsauthor{ Haoming Yang \And Ali Hasan \And  Vahid Tarokh }

\aistatsaddress{ Duke University \And  Morgan Stanley \And Duke University } ]

\begin{abstract}
Regularizing continual learning techniques is important for anticipating algorithmic behavior under new realizations of data. 
We introduce a new approach to continual learning by imposing the properties of a parabolic partial differential equation (PDE) to regularize the expected behavior of the loss over time.
This class of parabolic PDEs has a number of favorable properties that allow us to analyze the error incurred through forgetting and the error induced through generalization.
Specifically, we do this through imposing boundary conditions where the boundary is given by a memory buffer.
By using the memory buffer as a boundary, we can enforce long term dependencies by bounding the expected error by the boundary loss.
Finally, we illustrate the empirical performance of the method on a series of continual learning tasks. 
\end{abstract}

\section{INTRODUCTION}
Continual learning is important for ensuring algorithms adjust appropriately to new distributions of observed data. 
In many applications, rapidly adapting to changing environments and updating model behavior is necessary to ensure effective performance of a machine learning algorithm~\citep{wang2024comprehensive}.
However, questions regarding \emph{how} to adapt an algorithm to incoming data or changes in data distribution remain unsolved. 
For example, when observing samples that are possibly noisy or corrupted, these samples may be less important for model adaptation.
Similarly, previously observed samples may be necessary to maintain model performance on earlier observed tasks \citep{peng2023ideal}. 
These dynamics create a complex interplay between the newly observed data -- data that the model should adapt to and the previously observed data -- data that the model should maintain good performance on.
Designing a mathematical framework that encapsulates these behaviors -- specifically the expected learning behavior of the model -- is necessary for developing reliable continual learners. 
In this work, we describe a framework based on solutions to specific types of partial differential equations (PDEs) known as parabolic PDEs. 
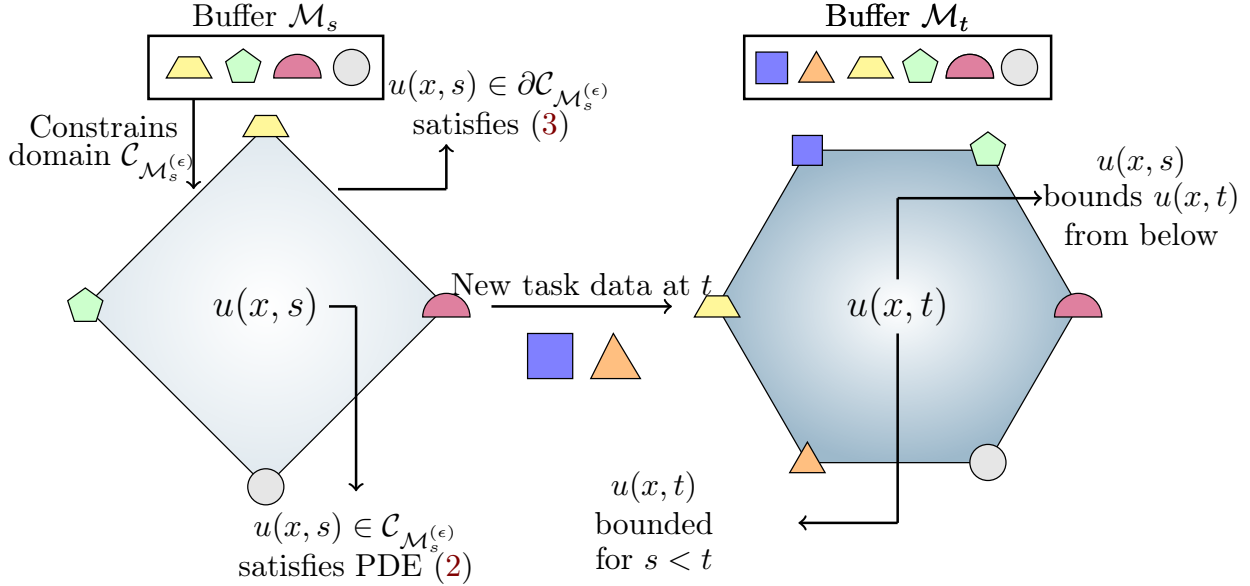
\begin{figure*}[]
\centering
\scalebox{1.2}{
    
    \begin{tikzpicture}
\node[shape=diamond, even odd rule,inner color=white,outer color=airforceblue!20, draw, minimum size=4cm]at (-3.5,0){};
\node[shape=trapezium, fill=yellow!50, draw]at (-3.5,2){};
\node[regular polygon, regular polygon sides=5, draw, fill=green!20] at (-5.5,0) {};
\node[shape=semicircle, fill=purple!50, draw]at (-1.5,0){};
\draw[fill=gray!20] (-3.5,-2) circle (0.2);

\node[shape=regular polygon, regular polygon sides=6, even odd rule,inner color=white,outer color=airforceblue!60, draw, minimum size=4cm]at (3.5,0){};
\node at (3.5,3.2) {Buffer $\mathcal{M}_t$};
\node[regular polygon, regular polygon sides=4, draw, fill=blue!50] at (2.5,1.73) {};
\node[shape=semicircle, fill=purple!50, draw]at (5.5,0){};
\node[regular polygon, regular polygon sides=3, draw, fill=orange!50,  inner sep=0.08cm] at (2.5,-1.73) {};
\node[shape=trapezium, fill=yellow!50, draw]at (1.5,-0){};
\node[regular polygon, regular polygon sides=5, draw, fill=green!20] at (4.5,1.73) {};
\node[shape=semicircle, fill=purple!50, draw]at (4.3,2.65){};
\draw[fill=gray!20] (4.5,-1.73) circle (0.2);

\draw[->,thick] (-1,0) -- (1,0) node[midway,above] {New task data at $t$};
\draw[fill=blue!50] (-0.6,-0.3) rectangle (-0.1,-0.8);
\draw[fill=orange!50] (0.35,-0.3) -- (0.1,-0.8) -- (0.65,-0.8) -- cycle;

\draw[thick] (-4.8,3) rectangle (-2.2,2.3);
\node at (-3.5,3.2) {Buffer $\mathcal{M}_s$};

\node[shape=trapezium, fill=yellow!50, draw]at (-4.35,2.65){};
\node[regular polygon, regular polygon sides=5, draw, fill=green!20] at (-3.75,2.65) {};
\node[shape=semicircle, fill=purple!50, draw]at (-3.15,2.65){};
\draw[fill=gray!20] (-2.55,2.65) circle (0.2);

\draw[thick] (5.2,3) rectangle (1.8,2.3);
\node at (3.5,3.2) {Buffer $\mathcal{M}_t$};
\draw[fill=blue!50] (1.93,2.47) rectangle (2.28,2.82);
\node[regular polygon, regular polygon sides=3, draw, fill=orange!50,  inner sep=0.08cm] at (2.6,2.6) {};
\node[shape=trapezium, fill=yellow!50, draw]at (3.2,2.65){};
\node[regular polygon, regular polygon sides=5, draw, fill=green!20] at (3.75,2.65) {};
\node[shape=semicircle, fill=purple!50, draw]at (4.3,2.65){};
\draw[fill=gray!20] (4.85,2.65) circle (0.2);

\node at (-5.3,2.0) {Constrains};
\node at (-5.3,1.6) {domain $\mathcal{C}_{\mathcal{M}_s^{(\epsilon)}}$};
\draw[<-,thick] (-4.3,1.3) -- (-4.3,2.3);
\node at (-1,2.4) {$\;\; u(x,s) \in \partial \mathcal{C}_{\mathcal{M}_s^{(\epsilon)}}$};
\node at (-1,2) {satisfies~\eqref{eq:bc}};

\draw[-,thick] (-2.7,1.3) -- (-1.5,1.3);
\draw[<-,thick] (-1.5,1.8) -- (-1.5,1.3);

\node at (-3.5,0) {\large$u(x,s)$};
\draw[-,thick] (-2.8,0) -- (-2.5,0);
\draw[->,thick] (-2.5,0) -- (-2.5,-2.05);
\node at (-2.5,-2.5) {$u(x, s) \in \mathcal{C}_{\mathcal{M}_s^{(\epsilon)}}$};
\node at (-2.5,-2.9) {satisfies PDE~\eqref{eq:pde}};

\node at (3.5,0) {\large $u(x,t)$};
\draw[-,thick] (3.5,-0.3) -- (3.5,-2.4);
\draw[->,thick] (3.5,-2.4) -- (2.4,-2.4);
\node at (0.8,-2.) {$u(x,t)$};
\node at (0.8, -2.4) {bounded};
\node at (0.8, -2.8) {for $s < t$};
\draw[-,thick] (3.5,0.3) -- (3.5,1.2);
\draw[->,thick] (3.5,1.2) -- (5.1,1.2);
\node at (6.2,1.6) {$u(x,s)$};
\node at (6.2, 1.2) {bounds $u(x,t)$};
\node at (6.2, 0.8) {from below};
\end{tikzpicture}}
    \caption{Properties of parabolic continual learner. At time $s$,  samples in the buffer $\mathcal{M}_s$ are evaluated to construct the boundary of expected loss profile $u(x,s)$. For $t>s$ a new task is introduced, PCL enforces the expected loss $u$ to satisfy the PDE in~\eqref{eq:pde} with boundary condition in~\eqref{eq:bc}. As long as $u(x,t)$ continues to satisfy the PDE, the loss profile of PCL can be anticipated for $t > s$.}
    \label{fig:fig1}
\end{figure*}
Parabolic PDEs are widely applied in machine learning, and it describe the rates of change over time~\citep{evans2022partial}.
Perhaps its most popular application is in diffusion based generative modeling, where the probability density linking the initial condition to target distribution is the solution to a parabolic PDE known as the Fokker-Planck equation~\citep{huang2021variational, xu2024base}.
Additionally, principles from stochastic control, which relate to the Hamilton-Jacobi-Bellman equation, are used in reinforcement learning problems~\citep{wiltzer2022distributional}.
In both of these cases, the parabolic PDE provides a structured way of relating quantities at different points in time to each other.
In our case, the PDE serves as the basis for regulating the expected loss of our proposed continual learner at different points in time.

As overarching goals, we are interested in designing a continual learner that satisfies two main properties: 
{\bf i) knowledge retention} -- the learner should maintain knowledge of previous tasks; 
{\bf ii) generalization} -- the learner should calibrate to new data observations and distributions.
To do this, we will impose properties on the evolution of the loss function such that we may be able to predict the behavior of the learner on new tasks. 
For {\bf i)}, the learner should be able to control the rate at which information from previous data is retained and lost through the growth of the loss function at previous times.
For {\bf ii)}, the rate at which the loss accumulates for any new data within a predefined region of the data space for future time points should be constrained.
Specifically, we will require that the loss satisfies a parabolic PDE and use the properties of this PDE to enforce properties i) and ii). 
These PDEs have also been well studied and a variety of favorable qualitative properties of their solutions exist. 
We will describe these as they pertain to the continual learning problem, but we note that properties of different PDEs has already been applied to other learning problems such as learning under distribution shifts \citep{hasan2025elliptic}.

\paragraph{Related Work}

Numerous continual learning methods have been proposed to adapt models to new training data. 
One major approach considers imposing a prior on the structure of the incoming data by imposing a level of regularization. 
These methods aim to retain the important weights for the previous tasks while optimizing the other parameters for the new task~\citep{li2017learning}, \citep{kirkpatrick2017overcoming}, \citep{zenke2017continual}.  
Another line of work maintains a memory buffer throughout the training process, where a small set of data from past tasks are saved to a buffer and trained together with the new tasks to reduce forgetting.
These works include~\cite{lopez2017gradient}, \cite{rolnick2019experience}, \cite{buzzega2020dark}. 
Follow-up work considered modifying the buffer or the sampling scheme to investigate the impact on learning. 
\citet{chrysakis2023online} considers a bias correction to the reservoir sampling algorithm. 
\citet{liu2022navigating} establishes a pseudo-task simulation to carefully choose the elements that are selected into the buffer. 
To further improve experience replay, saliency features were also included in the training process to guide the continual learner to look for similarities within data. 
Other works considered meta-learning based approaches to continual learning \citep{riemer2018learning, krishnan2020meta, gupta2020look, wu2024metacont}. 
\citet{wu2024metacont} established a link between regularization and meta-learning approaches through the second-order Hessian of loss. 
Finally, control approaches stemming from the Hamilton-Jacobi-Bellman equations were studied to connect the dynamic of the value function (or loss) to the changes in the parameters~\citep{krishnan2020meta}. 

Our work takes a different perspective, which we illustrate in Figure~\ref{fig:fig1}.
The proposed method shares similar ideas to PDE constraint regularization problems that is explore in robust learning under distribution shifts~\citep{hasan2025elliptic}.
We present a unified framework between regularization and replay approaches through constraining the expected loss over a boundary given by the replay buffer.
We do this by carefully choosing a sampling strategy such that the expected loss satisfies the solution to a parabolic PDE.
This allows us to use tools from the analysis of PDEs to understand the learning behavior of our algorithm. 
To summarize, our contributions are as follows:
First, we propose a framework for continual learning based on parabolic PDEs; then, we theoretically describe the favorable properties of this continual learner using the theory of parabolic PDEs; finally, we describe a computationally efficient method for implementing this loss.
We conclude the work by illustrating the performance of the parabolic continual learner on common continual learning benchmarks. 
 
\section{PARABOLIC CONTINUAL LEARNER}

To introduce the parabolic perspective on continual learning, we will first set up the continual learning problem.
We assume we observe data points $\{x_{t_i} \in \mathcal{X}_{t_i}\}$ at different time instances $t_i \in [0, T]$ associated with the $i^\text{th}$ task.  
Note that in the continual learning case, the task index $i$ is not directly observed; rather, data arrive with distributions related to the task. 
We define an \textit{non-negative} objective function for a given estimator $f_\theta$ with parameters $\theta$ as $\ell_{f_\theta} : \mathcal{X}  \to \mathbb{R}_+ $ that we will try to minimize for all $t \in [0, T]$.
For example, in a regression setting the loss may be $\ell_{f_\theta} := \|f_\theta(x_{t_i, 0}) - x_{t_i, 1} \|_2^2$ where $x_{t_i,j}$ is the $j^\text{th}$ component of the vector $x$.

\subsection{Intuition Behind Parabolic PDEs}
\label{sec:heat}
As mentioned, parabolic PDEs are used in a number of disciplines for defining the temporal evolution of different phenomena. 
The most well-known parabolic PDE is the heat equation, which describes the flow of temperature across a domain over time.
Here we will provide an intuitive example in the physical domain of parabolic PDEs to motivate our use in continual learning. 
Given a domain $[0, T] \times \mathcal{X} \subset \mathbb{R}_+ \times \mathbb{R}^d$, denoting the Laplacian by $\nabla^2$, we can define the heat equation as the following
$$
\underbrace{\frac{\partial u}{\partial t}}_{\textcircled{1}} = \underbrace{\vphantom{\frac{\partial u}{\partial t}}\nabla ^2 u(x)}_{\textcircled{2}} + \underbrace{\vphantom{\frac{\partial u}{\partial t}}g(x)}_{\textcircled{3}} \quad on \quad x,t \in \mathcal{X} \times [0,T]. 
$$
In this example, the function $u$ represents a temperature value and $\textcircled{1}$ describes the rate at which the temperature changes over time.
The term $\textcircled{2}$ states that the temperature changes are dominated by diffusive dynamics. The term 
$\textcircled{3}$ states that heat is added or removed according to the value of $g(x)$.
The regularity induced by solutions to equations of this type has a variety of useful properties.
In the case of continual learning, we will apply these concepts to impose a constraint on the learning dynamics through the parabolic. 
We refer the interested reader to~\citet{pardoux2014stochastic} for additional details and properties on this class of PDEs. For notation simplicity, we will denote $x,t \in \mathcal{X} \times [0,T]$ as $\mathcal{X}_t$ from now on. 

The overall approach of the proposed method is to define how $\ell_{f_\theta}$ changes over the full space of $\mathcal{X}$.
Consider a loss $\ell$ as defined above and an estimator $f_\theta$.

We define the \emph{continual loss profile} as the value of 
\begin{equation}
u(x,t) = \mathbb{E}[\ell_{f_\theta}(x_t)],
\label{eq:loss_prof}
\end{equation}
with the expectation taken with respect to some distribution of $x_t$ at time $t.$
Following most methods in continual learning, we will also assume we have access to a memory buffer $\mathcal{M}$ of fixed size $|\mathcal{M}|$. 
For each iteration $t$ in our training algorithm, we will solve the following optimization problem
$$
\min_{\theta_t} \mathbb{E}_{x\sim P(\mathcal{M}_t)} \left[ \ell_{f_{\theta_t}}(x) \right]
$$
where $x$ are sampled from distribution $P(\mathcal{M}_t)$ with support of $\mathcal{M}_t$. 
At each iteration, $\mathcal{M}_t$ is updated according to a sampling scheme~\citep{vitter1985random}. 
To investigate how $u$ changes over the domain over time, we will study its evolution $\frac{\partial u}{\partial t}$. 
In terms of the criteria i) and ii) $\frac{\partial u}{\partial t}$ should maintain long term dependencies on previous task realizations and be sufficiently adaptive to new tasks.
This means that $u(x, s)$ should never be too large for $s < t$ (to prevent forgetting), as well as $\partial_t u$, should be well-behaved for generalization for $t \to \infty$. This motivates the following model:
\begin{align}
\label{eq:pde}
    &\frac{\partial u}{\partial t} = \sigma \nabla^2 u(x, t)   + \ell_{f_\theta}(x) \quad on  \quad x,t \in \mathcal{C}_{\mathcal{M}_t^{(\epsilon)}}  / \mathcal{M}_t^{(\epsilon)},\\
    &u(x, t) = \ell_{f_\theta}(x)  \quad on  \quad x, t \in \mathcal{M}_t^{(\epsilon)}
\label{eq:bc}
\end{align}
In the above $\mathcal{M}^{(\epsilon)}$ denotes the $\epsilon$-ball expansion of the countable set in $\mathcal{M}$ given by $ \mathcal{M}^{(\epsilon)} = \cup_{x^{(i)} \in \mathcal{M}}\{ x \mid \|x - x^{(i)} \|_2^2 \leq \epsilon \}$ in order to provide a meaningful boundary, and $\mathcal{C}_\mathcal{M} \subset \mathcal{X}$ denotes the convex hull of $\mathcal{M}$ and provides a compact space to solve the learning problem over. 
In~\eqref{eq:pde} we first specify that the expected loss should primarily follow diffusive dynamics with sources coming from the loss at different points in the domain.  
In~\eqref{eq:bc} we specify that the expected loss should correspond to the observed data loss within the $\epsilon$ expanded memory buffer $\mathcal{M}^{(\epsilon)}$. Similar to notation before, $\mathcal{M}^{(\epsilon)}_t$ denotes the updated $\epsilon$ expanded memory buffer at each training iteration $t$, which lives on the domain $\mathcal{M}^{(\epsilon)} \times [0,T]$

\subsection{Interpretation of the Equation}
Relating equations~\eqref{eq:pde} and~\eqref{eq:bc} back to the interpretation in Section~\ref{sec:heat}, the value of the loss acts as energy added to the system.
The memory buffer $\mathcal{M}$ acts as the boundary condition where the initial energy is input into the system.
Losses that are large on the boundary will have large magnitude in the continual loss profile.
On the interior of the domain, energy is added through the loss value for unseen points in the convex hull of the memory buffer. 
Thinking of this in terms of optimization, the loss should be minimized for all points on the defined domain, namely in the convex hull of the memory buffer. 
We note that alternative definitions of this domain is also possible.
For example, one could solve the PDE on regions where the data are most likely to be found or where data are shifting towards. 
This provides an interdependence between the domain through the boundary conditions and the values of the loss profile.  

In our framework, the time variable $t$ corresponds to the arrival time of a new task. 

\subsection{Lower Order Terms}
\label{sec:first_order}
In practice, it may be beneficial to further enrich the model by adding first order derivative terms to the PDE.  Specifically, we rewrite~\eqref{eq:pde} as
\begin{equation}
    \frac{\partial u}{\partial t} = \sigma \nabla^2 u(x, t) + \mu(x)^\top \nabla u(x, t)   + \ell_{f_\theta}(x)
    \label{eq:parabolic}
\end{equation}
for $(x, t) \in \mathcal{C}_{\mathcal{M}_t^{(\epsilon)}}  / \mathcal{M}_t^{(\epsilon)}.$
The first order term provides a transport-type of interpretation where the energy is transported in a particular direction driven by the function $\mu(x) : \mathbb{R}^d \to \mathbb{R}^d$.
This can be useful in the learning setting by, for example, emphasizing different parts of the domain as specified by $\mu(x)$. In other words,
$\mu(x)$ can act as prior knowledge about which points contribute more weight to the loss. 
From the perspective of data arriving in an online setting, $\mu(x)$ could also be chosen to closely match the distribution of data arrivals.
This would bias the loss such that points within the trajectory satisfying $\mu(x)$ are provided with higher weights. 

\section{BOUNDS ON THE EXPECTED ERROR}
By following the parabolic PDE for the continual learning loss, we can analyze the theoretical properties of the continual loss profile. 
Most importantly, these allow us to analyze the expected error under this loss. Due to space constraints, the proofs for the proposition, corollary, and remarks in this section is deferred to Appendix \ref{appendix:proof}.

\subsection{Expected Forgetting Error}
Since we are explicitly constructing our domain by using the memory buffer $\mathcal{M}$, we can use the points within the buffer to describe what the expected error is for future points in time.
We consider this error the \emph{forgetting error} since this has to do with how the network forgets properties of data as new data points arrive. 
The first behavior of our continual learning landscape we discuss is the 
\begin{proposition}[Upper bound on expected forgetting error]
\label{prop:forget}
    Consider $u(x_{t-\tau},t-\tau)$ for $x_{t-\tau} \in \mathcal{C}_{\mathcal{M}_t}$ and $\tau \geq 0$.
    Additionally suppose that the learning landscape satisfies the following parabolic PDE $$\frac{\partial u}{\partial t} = -\nabla^2 u(x,t) + \ell_{f_\theta}(x) \quad on \quad (x, t) \in \mathcal{C}_{\mathcal{M}_t^{(\epsilon)}}$$
    Then the following inequality holds: $$u(x, t-\tau) \leq \max_{x, s \in \mathcal{M}_t^{(\epsilon)} \cup \partial \mathcal{C}_{\mathcal{M}_t^{(\epsilon)}} } u(x, s).$$
\end{proposition}
Proposition~\ref{prop:forget} allows us to understand how the loss of previous tasks grows as the learner changes in time. 
In other words, given the memory buffer $\mathcal{M}_t^{(\epsilon)}$ at time $t$, the expected error over all tasks within the convex hull of $\mathcal{M}_t^{(\epsilon)}$ is bounded by the error at the current time. 
This provides a unified perspective on the data in the memory buffer and the expected loss for previous time intervals.

\subsection{Expected Generalization Error}
The second component we wanted to enforce was the generalization of the algorithm to new data points. 
Suppose we observe our last data point at time $T$, we define the generalization error as the expected error for any time $s > T$.
As a corollary to Proposition~\ref{prop:forget}, we can reverse the inequalities and quantify a lower bound on the error for future points in time:
\begin{corollary}[Lower bound on expected generalization error]
\label{corr:general}
    Consider $u(x_{t+\tau},t+\tau)$ for $x_{t+\tau} \in \mathcal{C}_{\mathcal{M}_t}$ and $\tau \geq 0$. Suppose that the expected continual learning loss $u$ satisfies the following parabolic PDE
    $$\frac{\partial u}{\partial t} = \nabla^2 u(x,t) 
    + \ell_{f_\theta}(x) \quad on \quad (x, t) \in \mathcal{C}_{\mathcal{M}^{(\epsilon)}} \times [T, \infty)$$
    and that the function $\ell_{f_\theta}(x)$ is $C-$Lipshitz in $x$. 
    Then the following inequality holds:
    $$\min_{x, s \in \mathcal{M}_t \cup \partial \mathcal{C}_{\mathcal{M}^{(\epsilon)}_T} } u(x, s) \leq u(x, t+\tau) \leq  C\tau + \max_{ x \in \mathcal{C}_{\mathcal{M}^{(\epsilon)}_T}} u(x,s) .$$
\end{corollary}
where $\mathcal{M}_T$ is the memory buffer at time $T$. Corollary~\ref{corr:general} again uses the construction of the memory buffer to bound the expected error for new tasks occurring in the future.

\subsection{Influence of First-Order Terms}
Finally, we will study what happens when we include the first order derivative terms in the PDE.
Consider the continual learning setting where we have prior knowledge of a distribution from which we would like to sample. 
We will assume that this is a distribution parameterized by a convex function $\varphi$ and the density is given by $p(x) \propto \exp(\varphi(x))$. 
The following remark describes how the upper bound changes when adding the loss. 
\begin{remark}[An Upper Bound on the Negative Log Loss]
Suppose we optimize the continual loss profile under the following PDE
\begin{equation}
\frac{\partial u^\varphi }{\partial t} = \nabla \varphi(x) ^ \top \nabla u^\varphi (x) + \frac12 \nabla^2 u^\varphi (x) + \ell_{f_\theta}(x)
\label{eq:pde_first}
\end{equation}
where $\varphi(x)$ is convex with minimum at $x^\star$. 
Then the negative log loss of the solution $u^\varphi $ for any given $\ell_{f_\theta}$ is bounded from above by a value greater than or equal to the upper bound of the solution to~\eqref{eq:pde} under the same $\ell_{f_\theta}$.
That is, optimizing~\eqref{eq:pde_first} has a greater upper bound than the PDE without the linear term for all points except $x^\star$ where the bounds are equal.
\end{remark}
From this remark, the upper bound of the expected loss for all points except $x^\star$ is larger than the upper bound without the first order terms, suggesting an emphasis on the region outside of $x^\star$.

\section{COMPUTING THE PARABOLIC LOSS}

Solving high dimensional PDEs such as~\eqref{eq:pde} is difficult using traditional methods such as finite differences or finite elements. 
Fortunately, there exists a connection between stochastic processes, which we can easily sample, and solutions to PDEs of the type in~\eqref{eq:pde}. 
We will first describe this connection and then describe the algorithms which we use to impose the necessary constraints. 
\begin{algorithm}[ht]
\caption{Parabolic continual training to solve PDE described in \eqref{eq:pde}}
\label{alg:pcl}
\begin{algorithmic}
\STATE {Input: Data $X_{\text{all}},y_{\text{all}}$, related Brownian bridge hyperparameters.}
\STATE {Initialize: neural network $f_\theta$, buffer $\mathcal{M}$.}
\FOR{$X,y$ in mini-batched $X_{\text{all}},y_{\text{all}}$}
     \IF{$\mathcal{M}$ is not empty}
     \STATE Sample from $\mathcal{M}$, obtain $X_\mathcal{M}, y_\mathcal{M}$.
     \STATE $X = \operatorname{Concatenate}(X,X_\mathcal{M})$.
     \STATE $y = \operatorname{Concatenate}(y,y_\mathcal{M})$.
     \ENDIF
     \STATE Obtained $X',y'$ as Brownian bridge terminal condition by shuffling $X,y$.
     \STATE Sample Brownian bridges $X_s, y_s \sim \mathrm{BB}_{X,y}^{X',y'}$ for arbitrary timestep $s$ (see Algorithm~\ref{alg:bbridge}).
     \STATE Compute loss $\ell$ by integrating $\ell(f(X_s),y_s)$ using the Euler's method over $s \in [0,1]$.
     \STATE Optimize $\ell$ using gradient-based optimizer
     \STATE Update Buffer $\mathcal{M}$ with reservoir sampling.
\ENDFOR
\end{algorithmic}
\end{algorithm}

\subsection{Feynman-Kac Formula}

The Feynman-Kac formula bridges the gap between SDEs and PDEs. 
The main idea is to use the representation of a PDE in terms of its SDE to provide a computationally scalable technique for solving the PDE through Monte Carlo methods.
We refer to~\citet{pardoux2014stochastic} for a more thorough treatment of the connection.
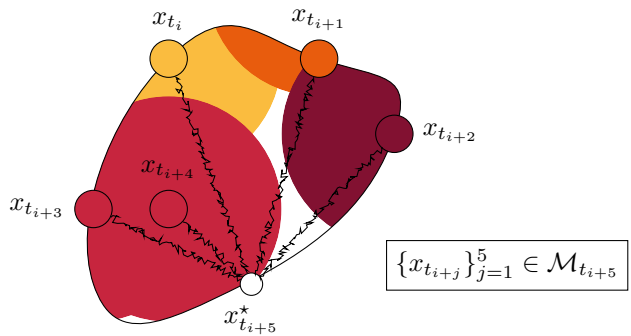
\begin{figure}
\scalebox{1.0}{
       \begin{tikzpicture}
    \begin{scope}
    \clip plot [smooth cycle, tension=1] coordinates {(0,-1) (1,2) (3,2) (4,1) (2,-1)};
    
    \fill[path fading=fade out, color=pastelPink] (1,2) circle (1.5);
    \fill[path fading=fade out, color=pastelBlue] (3,3) circle (1.5);
    \fill[path fading=fade out, color=pastelYellow] (4,1) circle (1.5);
    \fill[path fading=fade out, color=pastelGreen] (0,0) circle (1.5);
    \fill[path fading=fade out, color=pastelGreen] (1,0) circle (1.5);
    \end{scope}

    \draw plot [smooth cycle, tension=1] coordinates {(0,-1) (1,2) (3,2) (4,1) (2,-1)};
    
    \node[draw=black, fill=white, circle, inner sep=3pt, label=below:$x^\star_{t_{i+5}}$] (interest) at (2.1,-1.0) {};
    
    \node[draw=black,fill=pastelPink, circle, inner sep=5pt, label=above:$x_{t_i}$] (bc1) at (1,2) {};
    \node[draw=black,fill=pastelBlue, circle, inner sep=5pt, label=above:$x_{t_{i+1}}$] (bc2) at (3,2) {};
    \node[draw=black,fill=pastelYellow, circle, inner sep=5pt, label=right:$x_{t_{i+2}}$] (bc3) at (4,1) {};
    \node[draw=black,fill=pastelGreen, circle, inner sep=5pt, label=left:$x_{t_{i+3}}$] (bc4) at (0,0) {};
    \node[draw=black,fill=pastelGreen, circle, inner sep=5pt, label=above:$x_{t_{i+4}}$] (bc5) at (1,0) {};

    \draw[brownian] (interest) -- (bc1);

    \draw[brownian] (interest) -- (bc2);
    \draw[brownian] (interest) -- (bc3);
    \draw[brownian] (interest) -- (bc4);
    \draw[brownian] (interest) -- (bc5);

    \node[draw] at (5.5,-0.75) {$\{x_{t_{i+j}} \}_{j=1}^5\in \mathcal{M}_{t_{i+5}}$};

\end{tikzpicture}
}
\caption{Memory buffer $\mathcal{M}_t$ changes for different time steps and dictates the boundary on which the solution is found.} 
\label{fig:fk_cartoon}

\end{figure}
The main idea is that the PDE in~\eqref{eq:pde} can be solved by the following expectation:
\begin{align}
\nonumber &u(x, t) =  \\
&\mathbb{E} \left [ \ell_{f_\theta}(X_{t \wedge \tau_{\mathcal{M}^{(\epsilon)}}} ) + \int_0^{t\wedge \tau_{\mathcal{M}^{(\epsilon)}}} \ell_{f_\theta}(X_s) \mathrm{d}s \mid X_0 = x\right ]
    \label{eq:fk}
\end{align}
\begin{algorithm}[t!]
\caption{Sampling a Brownian bridge}
\label{alg:bbridge}
\begin{algorithmic}
\STATE {Input: Initial condition $X \in \mathbb{R}^d$, terminal condition $X' \in \mathbb{R}^d$, diffusion coefficient $\sigma$, number of time steps $k$}
\STATE {Let $\Delta_t = \frac{T- t_0}{k}$ be the time resolution, where $t_0 = 0, t_k = 1$, and $t_{i+1} - t_{i} = \Delta_t$}
\STATE {Sample $k$ samples from standard normal $N(0,I), I = \mathds{1}^{d}$, denoted as $\Delta W_t$}
\STATE {Integrate samples to form Brownian motion $W_t = \sum_{s=0}^t \sigma \Delta W_s \sqrt{\Delta_t}$}
\STATE {Set $W_0= 0$; then compute $W_t = W_t + X$}
\STATE {Set Brownian bridge as $BB = W - t(W_k - X')$}
\end{algorithmic}
\end{algorithm}
where $\tau_{\mathcal{M}^{(\epsilon)}} = \inf_{s > 0} X_s \in \mathcal{M}^{(\epsilon)}$, i.e. the first hitting time of $X_s$ in the $\mathcal{M}^{(\epsilon)}$ memory buffer and $\wedge$ is the minimum operator.
Now, all we need are samples of $X_s$ and we can compute $u$ and minimize~\eqref{eq:fk} as a function of $\theta$.  
We can also include first order derivatives as in Section~\ref{sec:first_order} by computing the Radon-Nikodym derivative:
\begin{align*}
\frac{\mathrm{d}P_\mu}{\mathrm{d}Q} := \exp \bigg (\int_0^{t\wedge \tau_{\mathcal{M}^{(\epsilon)}}} & \mu^\top(X_s) \mathrm{d} X_s \\ &- \frac12 \int_0^{t\wedge \tau_{\mathcal{M}^{(\epsilon)}}}\mu^\top \mu(X_s) \mathrm{d}s \bigg ).
\end{align*}
Minimizing the expectation in $u_\mu(x,t) = \mathbb{E} \left [ \left(\ell_{f_\theta}(X_{t \wedge \tau_{\mathcal{M}^{(\epsilon)}}} ) + \int_0^{t\wedge \tau_{\mathcal{M}^{(\epsilon)}}} \ell_{f_\theta}(X_s) \mathrm{d}s \right) \frac{\mathrm{d}P_\mu}{\mathrm{d}Q}\mid X_0 = x\right ]$ leads to solving the PDE in~\eqref{eq:pde_first}.

\subsection{Computational Simplifications}
To evaluate the expectations in~\eqref{eq:fk}, we use an approximation to prevent long integration times.
Specifically, we compute Brownian bridges between points in the memory buffer and the incoming data points. 
These are then used in Algorithm~\ref{alg:pcl} to minimize the loss according to the solution of the PDE. 
Using the Brownian bridges allows us to compute the error over regions of the data space that are most likely to be visited according to the observed data distribution. 
We present the algorithm to sample a Brownian bridge in Algorithm~\ref{alg:bbridge}.
Moreover, this has the effect of choosing the first order derivative terms such that they are biased towards the interior of the convex hull of the data points. We note that this simplifcation results in algorithmic similarities to MixupER \citep{lim2024mixer}. However, mixup and Brownian bridges have fundamentally different qualities since Brownian bridges induces a diffusive property and thus solves the parabolic PDE. In the mixup case with no diffusion, some regions of the domain will never be sampled and the qualitative properties of parabolic PDEs do not hold. Thus, we will show that PCL outperforms mixupER empirically due to the favorable diffusive properties of Brownian bridges that satisfy the parabolic PDE. 
\begin{table*}[htbp]
    \centering
    \caption{Sequential CIFAR10, sequential CIFAR100 and sequential TinyImageNet at buffer size $|\mathcal{M}| = 1000$. Methods with the best average performance are \textbf{bolded}, and the second best are \underline{underlined}.} 
    \begin{tabular}{@{}llllllll@{}}
    \toprule
          & \multicolumn{2}{c}{Seq-CIFAR10} & \multicolumn{2}{c}{Seq-CIFAR100} & \multicolumn{2}{c}{Seq-TinyImageNet}    \\ \toprule
          & AAA & Acc & AAA & Acc & AAA & Acc \\ \midrule
SGD       & 34.17 $_{\pm{ 2.5 }}$ & 15.57 $_{\pm{ 1.3 }}$ & 7.91 $_{\pm{ 0.3 }}$ & 3.99 $_{\pm{ 0.5 }}$ & 5.05 $_{\pm{ 0.3 }}$ & 1.53 $_{\pm{ 0.2 }}$\\
ER        & 56.44 $_{\pm{ 0.8 }}$ & 42.15 $_{\pm{ 1.3 }}$ & 20.28 $_{\pm{ 0.3 }}$ & 13.94 $_{\pm{ 1.2 }}$ & 15.27 $_{\pm{ 0.6 }}$ & 9.06 $_{\pm{ 0.7 }}$\\
SI        & 34.52 $_{\pm{ 2.2 }}$ & 16.8 $_{\pm{ 0.3 }}$ & 7.99 $_{\pm{ 0.6 }}$ & 3.54 $_{\pm{ 0.7 }}$ & 5.15 $_{\pm{ 0.4 }}$ & 1.3 $_{\pm{ 0.3 }}$\\
AGEM      & 35.58 $_{\pm{ 0.6 }}$ & 17.14 $_{\pm{ 0.7 }}$ & 8.13 $_{\pm{ 0.3 }}$ & 4.25 $_{\pm{ 0.4 }}$ & 5.48 $_{\pm{ 0.2 }}$ & 1.3 $_{\pm{ 0.1 }}$\\
GEM       & 35.21 $_{\pm{ 0.7 }}$ & 15.8 $_{\pm{ 1.1 }}$ & 8.95 $_{\pm{ 0.7 }}$ & 5.35 $_{\pm{ 1.1 }}$ & 5.86 $_{\pm{ 0.6 }}$ & 2.32 $_{\pm{ 0.7 }}$\\
DER       & 35.7 $_{\pm{ 3.3 }}$ & 16.83 $_{\pm{ 1.7 }}$ & 8.5 $_{\pm{ 0.4 }}$ & 4.06 $_{\pm{ 0.6 }}$ & 5.25 $_{\pm{ 0.3 }}$ & 1.73 $_{\pm{ 0.2 }}$\\
DER++     & 45.89 $_{\pm{ 2.9 }}$ & 32.79 $_{\pm{ 3.8 }}$ & 10.7 $_{\pm{ 1.0 }}$ & 6.34 $_{\pm{ 0.9 }}$ & 7.49 $_{\pm{ 0.3 }}$ & 3.75 $_{\pm{ 1.4 }}$\\
On-EWC    & 35.38 $_{\pm{ 1.9 }}$ & 15.6 $_{\pm{ 1.4 }}$ & 9.59 $_{\pm{ 1.0 }}$ & 3.98 $_{\pm{ 0.9 }}$ & 5.95 $_{\pm{ 0.2 }}$ & 1.48 $_{\pm{ 0.1 }}$\\
LWF       & 32.95 $_{\pm{ 2.1 }}$ & 15.66 $_{\pm{ 2.5 }}$ & 8.12 $_{\pm{ 0.6 }}$ & 4.69 $_{\pm{ 0.7 }}$ & 5.4 $_{\pm{ 0.4 }}$ & 1.93 $_{\pm{ 0.2 }}$\\
ER-OBC    & 58.78 $_{\pm{ 2.0 }}$ & 44.68 $_{\pm{ 3.6 }}$ & 24.1 $_{\pm{ 0.5 }}$ & 15.65 $_{\pm{ 0.9 }}$ & \underline{17.31} $_{\pm{ 0.7 }}$ & 7.83 $_{\pm{ 0.6 }}$\\
CLSER     & 63.86 $_{\pm{ 1.5 }}$ & \underline{53.86} $_{\pm{ 4.9 }}$ & 25.21 $_{\pm{ 1.0 }}$ & 17.7 $_{\pm{ 0.9 }}$ & 17.07 $_{\pm{ 0.3 }}$ & 8.83 $_{\pm{ 0.7 }}$\\
MixupER & 60.65 $_{\pm{ 2.2 }}$ & 47.77 $_{\pm{ 0.2 }}$ & 22.69 $_{\pm{ 0.9 }}$ & 17.96 $_{\pm{ 0.8 }}$ & 16.66 $_{\pm{ 0.3 }}$ & 9.54 $_{\pm{ 0.8 }}$\\
VRMCL     & \textbf{65.04} $_{\pm{ 0.9 }}$ & 52.81 $_{\pm{ 2.0 }}$ & \textbf{26.96} $_{\pm{ 0.6 }}$ & \underline{19.33} $_{\pm{ 0.8 }}$ & \textbf{20.05} $_{\pm{ 0.7 }}$ & \textbf{12.15} $_{\pm{ 0.7 }}$\\
PCL   & \underline{64.55} $_{\pm{ 1.9 }}$ & \textbf{54.98} $_{\pm{ 4.1 }}$ & \underline{25.82} $_{\pm{ 1.2 }}$ & \textbf{20.54} $_{\pm{ 0.9 }}$ & 16.99 $_{\pm{ 0.5 }}$ & \underline{10.12} $_{\pm{ 1.2 }}$\\
\bottomrule
    \end{tabular}
    \label{tab:seqallfixbuff}
\end{table*}

\begin{table*}[t!]
    \centering
    \caption{Sequential CIFAR-100 trained with buffer size $|\mathcal{M}| \in \{200,600,1000\}$. Methods with the best average performance are \textbf{bolded}, and the second best are \underline{underlined}.}
    \begin{tabular}{@{}llllllll@{}}
    \toprule
        & \multicolumn{2}{c}{$|\mathcal{M}|=200$} & \multicolumn{2}{c}{$|\mathcal{M}|=600$} & \multicolumn{2}{c}{$|\mathcal{M}|=1000$} \\ \toprule
        & AAA & Acc & AAA & Acc & AAA & Acc \\ \midrule
SGD     & 7.91 $_{\pm{ 0.3 }}$ & 3.99 $_{\pm{ 0.5 }}$ & 7.91 $_{\pm{ 0.3 }}$ & 3.99 $_{\pm{ 0.5 }}$ & 7.91 $_{\pm{ 0.3 }}$ & 3.99 $_{\pm{ 0.5 }}$\\
ER      & 16.25 $_{\pm{ 0.9 }}$ & 8.24 $_{\pm{ 0.3 }}$ & 19.12 $_{\pm{ 0.9 }}$ & 12.87 $_{\pm{ 1.4 }}$ & 20.28 $_{\pm{ 0.3 }}$ & 13.94 $_{\pm{ 1.2 }}$\\
SI      & 7.99 $_{\pm{ 0.6 }}$ & 3.54 $_{\pm{ 0.7 }}$ & 7.99 $_{\pm{ 0.6 }}$ & 3.54 $_{\pm{ 0.7 }}$ & 7.99 $_{\pm{ 0.6 }}$ & 3.54 $_{\pm{ 0.7 }}$\\
AGEM    & 7.96 $_{\pm{ 0.4 }}$ & 4.32 $_{\pm{ 0.4 }}$ & 7.89 $_{\pm{ 0.3 }}$ & 4.38 $_{\pm{ 0.4 }}$ & 8.13 $_{\pm{ 0.3 }}$ & 4.25 $_{\pm{ 0.4 }}$\\
GEM     & 8.87 $_{\pm{ 0.7 }}$ & 4.42 $_{\pm{ 0.6 }}$ & 9.17 $_{\pm{ 0.7 }}$ & 5.52 $_{\pm{ 0.4 }}$ & 8.95 $_{\pm{ 0.7 }}$ & 5.35 $_{\pm{ 1.1 }}$\\
DER     & 8.52 $_{\pm{ 0.6 }}$ & 3.85 $_{\pm{ 0.4 }}$ & 8.78 $_{\pm{ 0.3 }}$ & 3.78 $_{\pm{ 0.7 }}$ & 8.5 $_{\pm{ 0.4 }}$ & 4.06 $_{\pm{ 0.6 }}$\\
DER++   & 9.93 $_{\pm{ 0.6 }}$ & 5.02 $_{\pm{ 1.6 }}$ & 10.66 $_{\pm{ 0.9 }}$ & 5.72 $_{\pm{ 1.5 }}$ & 10.7 $_{\pm{ 1.0 }}$ & 6.34 $_{\pm{ 0.9 }}$\\
On-EWC  & 9.59 $_{\pm{ 1.0 }}$ & 3.98 $_{\pm{ 0.9 }}$ & 9.59 $_{\pm{ 1.0 }}$ & 3.98 $_{\pm{ 0.9 }}$ & 9.59 $_{\pm{ 1.0 }}$ & 3.98 $_{\pm{ 0.9 }}$\\
LWF     & 8.12 $_{\pm{ 0.6 }}$ & 4.69 $_{\pm{ 0.7 }}$ & 8.12 $_{\pm{ 0.6 }}$ & 4.69 $_{\pm{ 0.7 }}$ & 8.12 $_{\pm{ 0.6 }}$ & 4.69 $_{\pm{ 0.7 }}$\\
ER-OBC  & 17.22 $_{\pm{ 0.5 }}$ & 8.53 $_{\pm{ 0.2 }}$ & 22.06 $_{\pm{ 0.5 }}$ & 12.72 $_{\pm{ 1.1 }}$ & 24.1 $_{\pm{ 0.5 }}$ & 15.65 $_{\pm{ 0.9 }}$\\
CLSER   & 20.01 $_{\pm{ 0.4 }}$ & 10.56 $_{\pm{ 0.6 }}$ & 23.44 $_{\pm{ 1.2 }}$ & 14.72 $_{\pm{ 0.5 }}$ & 25.21 $_{\pm{ 1.0 }}$ & 17.7 $_{\pm{ 0.9 }}$\\
MixupER & 18.75 $_{\pm{ 0.7 }}$ & 11.14 $_{\pm{ 0.2 }}$ & 21.2 $_{\pm{ 1.4 }}$ & 15.35 $_{\pm{ 1.6 }}$ & 22.69 $_{\pm{ 0.9 }}$ & 17.96 $_{\pm{ 0.8 }}$ \\
VRMCL   & \textbf{22.33} $_{\pm{ 0.5 }}$ & \textbf{12.69} $_{\pm{ 0.8 }}$ & \textbf{25.75} $_{\pm{ 0.4 }}$ & \underline{17.5} $_{\pm{ 0.6 }}$ & \textbf{26.96} $_{\pm{ 0.6 }}$ & \underline{19.33} $_{\pm{ 0.8 }}$\\
PCL & \underline{20.51} $_{\pm{ 0.7 }}$ & \underline{12.52} $_{\pm{ 0.7 }}$ & \underline{24.62} $_{\pm{ 1.5 }}$ & \textbf{17.63} $_{\pm{ 1.6 }}$ & \underline{25.82} $_{\pm{ 1.2 }}$ & \textbf{20.54} $_{\pm{ 0.9 }}$\\
\bottomrule
    \end{tabular}
    \label{tab:seqcifar100diffbuff}
\end{table*}
\begin{table*}[t!]
    \centering
    \caption{Training with 50\% corrupted label, Sequential CIFAR10, sequential CIFAR100 and sequential TinyImageNet at buffer size $|\mathcal{M}| = 1000$. Methods with the best average performance are \textbf{bolded}, and the second best are \underline{underlined}.}
    \begin{tabular}{@{}llllllll@{}}
    \toprule
          & \multicolumn{2}{c}{Seq-CIFAR10} & \multicolumn{2}{c}{Seq-CIFAR100} & \multicolumn{2}{c}{Seq-TinyImageNet}    \\ \toprule
          & AAA & Acc & AAA & Acc & AAA & Acc \\ \midrule
SGD       & 30.6 $_{\pm{ 2.9 }}$ & 14.38 $_{\pm{ 1.3 }}$ & 5.36 $_{\pm{ 0.2 }}$ & 2.15 $_{\pm{ 0.7 }}$ & 3.24 $_{\pm{ 0.3 }}$ & 0.9 $_{\pm{ 0.2 }}$\\
ER        & 50.41 $_{\pm{ 2.6 }}$ & 37.03 $_{\pm{ 2.2 }}$ & 12.49 $_{\pm{ 0.6 }}$ & 6.78 $_{\pm{ 0.9 }}$ & 8.1 $_{\pm{ 0.5 }}$ & 3.33 $_{\pm{ 0.1 }}$\\
SI        & 30.5 $_{\pm{ 3.7 }}$ & 14.91 $_{\pm{ 0.7 }}$ & 5.3 $_{\pm{ 0.1 }}$ & 2.07 $_{\pm{ 0.6 }}$ & 3.17 $_{\pm{ 0.2 }}$ & 0.75 $_{\pm{ 0.1 }}$\\
AGEM      & 30.98 $_{\pm{ 3.0 }}$ & 14.45 $_{\pm{ 0.9 }}$ & 5.86 $_{\pm{ 0.2 }}$ & 2.66 $_{\pm{ 0.4 }}$ & 3.46 $_{\pm{ 0.3 }}$ & 0.88 $_{\pm{ 0.1 }}$\\
GEM       & 28.99 $_{\pm{ 3.2 }}$ & 14.96 $_{\pm{ 1.1 }}$ & 5.41 $_{\pm{ 0.2 }}$ & 2.52 $_{\pm{ 0.8 }}$ & 3.42 $_{\pm{ 0.2 }}$ & 1.03 $_{\pm{ 0.1 }}$\\
DER       & 31.95 $_{\pm{ 2.5 }}$ & 12.54 $_{\pm{ 1.0 }}$ & 5.7 $_{\pm{ 0.5 }}$ & 3.03 $_{\pm{ 0.1 }}$ & 3.82 $_{\pm{ 0.3 }}$ & 0.86 $_{\pm{ 0.1 }}$\\
DER++     & 44.15 $_{\pm{ 5.7 }}$ & 28.35 $_{\pm{ 5.1 }}$ & 6.45 $_{\pm{ 0.4 }}$ & 3.58 $_{\pm{ 0.3 }}$ & 4.44 $_{\pm{ 0.4 }}$ & 1.59 $_{\pm{ 0.1 }}$\\
On-EWC    & 28.0 $_{\pm{ 2.5 }}$ & 12.29 $_{\pm{ 3.1 }}$ & 5.74 $_{\pm{ 0.6 }}$ & 2.39 $_{\pm{ 0.5 }}$ & 3.28 $_{\pm{ 0.1 }}$ & 0.52 $_{\pm{ 0.1 }}$\\
LWF       & 26.66 $_{\pm{ 2.4 }}$ & 13.31 $_{\pm{ 2.7 }}$ & 5.92 $_{\pm{ 0.4 }}$ & 2.62 $_{\pm{ 1.0 }}$ & 3.52 $_{\pm{ 0.4 }}$ & 0.86 $_{\pm{ 0.2 }}$\\
ER-OBC    & 48.86 $_{\pm{ 2.2 }}$ & 34.08 $_{\pm{ 2.3 }}$ & 14.35 $_{\pm{ 0.9 }}$ & 8.06 $_{\pm{ 0.6 }}$ & 7.98 $_{\pm{ 0.4 }}$ & 2.32 $_{\pm{ 0.2 }}$\\
CLSER     & \underline{53.26} $_{\pm{ 2.1 }}$ & \underline{39.28} $_{\pm{ 3.3 }}$ & \textbf{15.6} $_{\pm{ 0.9 }}$ & 8.65 $_{\pm{ 0.6 }}$ & \textbf{9.97} $_{\pm{ 0.8 }}$ & \underline{3.65} $_{\pm{ 0.3 }}$\\
VRMCL     & 51.14 $_{\pm{ 2.3 }}$ & 37.67 $_{\pm{ 2.7 }}$ & \underline{15.16} $_{\pm{ 0.3 }}$ & \underline{8.88} $_{\pm{ 0.5 }}$ & 8.23 $_{\pm{ 0.5 }}$ & 3.31 $_{\pm{ 0.2 }}$\\
PCL   & \textbf{53.65} $_{\pm{ 2.7 }}$ & \textbf{41.86} $_{\pm{ 5.3 }}$ & 14.82 $_{\pm{ 0.6 }}$ & \textbf{9.53} $_{\pm{ 0.7 }}$ & \underline{9.96} $_{\pm{ 0.8 }}$ & \textbf{5.15} $_{\pm{ 0.6 }}$\\
\bottomrule
    \end{tabular}
    \label{tab:seqallfixbuff_corrupt}
\end{table*}
\begin{table*}[t!]
    \centering
    \caption{Training with 50\% corrupted label, Sequential CIFAR-100 trained with buffer size $|\mathcal{M}| \in \{200,600,1000\}$. Methods with the best average performance are \textbf{bolded}, and the second best are \underline{underlined}.}
    \begin{tabular}{@{}llllllll@{}}
    \toprule
        & \multicolumn{2}{c}{$|\mathcal{M}=200|$} & \multicolumn{2}{c}{$|\mathcal{M}=600|$} & \multicolumn{2}{c}{$|\mathcal{M}=1000|$} \\ \toprule
        & AAA & Acc & AAA & Acc & AAA & Acc \\ \midrule
SGD     & 5.36 $_{\pm{ 0.2 }}$ & 2.15 $_{\pm{ 0.7 }}$ & 5.36 $_{\pm{ 0.2 }}$ & 2.15 $_{\pm{ 0.7 }}$ & 5.36 $_{\pm{ 0.2 }}$ & 2.15 $_{\pm{ 0.7 }}$\\
ER      & 9.64 $_{\pm{ 1.1 }}$ & 5.29 $_{\pm{ 0.3 }}$ & 10.78 $_{\pm{ 0.4 }}$ & 6.76 $_{\pm{ 0.3 }}$ & 12.49 $_{\pm{ 0.6 }}$ & 6.78 $_{\pm{ 0.9 }}$\\
SI      & 5.3 $_{\pm{ 0.1 }}$ & 2.07 $_{\pm{ 0.6 }}$ & 5.3 $_{\pm{ 0.1 }}$ & 2.07 $_{\pm{ 0.6 }}$ & 5.3 $_{\pm{ 0.1 }}$ & 2.07 $_{\pm{ 0.6 }}$\\
AGEM    & 5.62 $_{\pm{ 0.3 }}$ & 2.91 $_{\pm{ 0.5 }}$ & 5.51 $_{\pm{ 0.3 }}$ & 2.41 $_{\pm{ 0.4 }}$ & 5.86 $_{\pm{ 0.2 }}$ & 2.66 $_{\pm{ 0.4 }}$\\
GEM     & 5.75 $_{\pm{ 0.2 }}$ & 3.17 $_{\pm{ 0.9 }}$ & 5.4 $_{\pm{ 0.3 }}$ & 2.61 $_{\pm{ 0.6 }}$ & 5.41 $_{\pm{ 0.2 }}$ & 2.52 $_{\pm{ 0.8 }}$\\
DER     & 5.67 $_{\pm{ 0.4 }}$ & 2.95 $_{\pm{ 0.3 }}$ & 5.84 $_{\pm{ 0.6 }}$ & 2.7 $_{\pm{ 0.3 }}$ & 5.7 $_{\pm{ 0.5 }}$ & 3.03 $_{\pm{ 0.1 }}$\\
DER++   & 6.75 $_{\pm{ 0.3 }}$ & 3.94 $_{\pm{ 0.7 }}$ & 6.72 $_{\pm{ 0.4 }}$ & 3.43 $_{\pm{ 0.5 }}$ & 6.45 $_{\pm{ 0.4 }}$ & 3.58 $_{\pm{ 0.3 }}$\\
On-EWC  & 5.74 $_{\pm{ 0.6 }}$ & 2.39 $_{\pm{ 0.5 }}$ & 5.74 $_{\pm{ 0.6 }}$ & 2.39 $_{\pm{ 0.5 }}$ & 5.74 $_{\pm{ 0.6 }}$ & 2.39 $_{\pm{ 0.5 }}$\\
LWF     & 5.92 $_{\pm{ 0.4 }}$ & 2.62 $_{\pm{ 1.0 }}$ & 5.92 $_{\pm{ 0.4 }}$ & 2.62 $_{\pm{ 1.0 }}$ & 5.92 $_{\pm{ 0.4 }}$ & 2.62 $_{\pm{ 1.0 }}$\\
ER-OBC  & 10.91 $_{\pm{ 0.5 }}$ & 5.25 $_{\pm{ 0.3 }}$ & 12.68 $_{\pm{ 0.9 }}$ & 6.85 $_{\pm{ 0.4 }}$ & 14.35 $_{\pm{ 0.9 }}$ & 8.06 $_{\pm{ 0.6 }}$\\
CLSER   & \underline{12.18} $_{\pm{ 0.8 }}$ & \underline{5.76} $_{\pm{ 0.2 }}$ & \textbf{14.21} $_{\pm{ 0.3 }}$ & 7.47 $_{\pm{ 0.3 }}$ & \textbf{15.6} $_{\pm{ 0.9 }}$ & 8.65 $_{\pm{ 0.6 }}$\\
VRMCL   & 11.32 $_{\pm{ 0.4 }}$ & 5.48 $_{\pm{ 0.9 }}$ & \underline{13.39} $_{\pm{ 0.4 }}$ & \underline{7.95} $_{\pm{ 0.5 }}$ & \underline{15.16} $_{\pm{ 0.3 }}$ & \underline{8.88} $_{\pm{ 0.5 }}$\\
PCL & \textbf{12.83} $_{\pm{ 0.5 }}$ & \textbf{7.05} $_{\pm{ 0.5 }}$ & 13.14 $_{\pm{ 1.1 }}$ & \textbf{8.42} $_{\pm{ 0.5 }}$ & 14.82 $_{\pm{ 0.6 }}$ & \textbf{9.53} $_{\pm{ 0.7 }}$\\
\bottomrule
    \end{tabular}
    \label{tab:seqallcifar100diffbuff_corrupt}
\end{table*}

\begin{table*}[ht]
    \centering
    \caption{Sequential CIFAR-10 trained with different imbalance settings. Methods with the best average performance are \textbf{bolded}, and the second best are \underline{underlined}.}
    \begin{tabular}{@{}llllllll@{}}
    \toprule
        & \multicolumn{2}{c}{$\gamma=2$ Normal} & \multicolumn{2}{c}{$\gamma=2$ Reversed} & \multicolumn{2}{c}{$\gamma=2$ Random} \\ \toprule
        & AAA & Acc & AAA & Acc & AAA & Acc \\ \midrule

SGD     & 36.79 $_{\pm{ 1.1 }}$ & 16.43 $_{\pm{ 1.2 }}$ & 36.94 $_{\pm{ 0.7 }}$ & 16.92 $_{\pm{ 0.3 }}$ & 36.48 $_{\pm{ 0.9 }}$ & 15.83 $_{\pm{ 1.1 }}$\\
ER      & 60.35 $_{\pm{ 2.2 }}$ & 47.91 $_{\pm{ 3.6 }}$ & 61.05 $_{\pm{ 2.6 }}$ & 49.2 $_{\pm{ 1.4 }}$ & 60.13 $_{\pm{ 2.4 }}$ & 47.04 $_{\pm{ 2.4 }}$\\
AGEM    & 35.99 $_{\pm{ 0.6 }}$ & 15.97 $_{\pm{ 1.8 }}$ & 35.14 $_{\pm{ 0.6 }}$ & 16.41 $_{\pm{ 0.8 }}$ & 36.39 $_{\pm{ 0.8 }}$ & 16.2 $_{\pm{ 1.4 }}$\\
GEM     & 36.56 $_{\pm{ 0.9 }}$ & 16.51 $_{\pm{ 0.9 }}$ & 35.51 $_{\pm{ 1.2 }}$ & 16.09 $_{\pm{ 0.7 }}$ & 36.33 $_{\pm{ 0.8 }}$ & 16.1 $_{\pm{ 0.8 }}$\\
DER     & 38.66 $_{\pm{ 2.7 }}$ & 16.48 $_{\pm{ 1.6 }}$ & 38.45 $_{\pm{ 1.7 }}$ & 17.63 $_{\pm{ 0.8 }}$ & 38.79 $_{\pm{ 2.7 }}$ & 17.61 $_{\pm{ 2.6 }}$\\
DER++   & 54.08 $_{\pm{ 2.8 }}$ & 41.75 $_{\pm{ 5.1 }}$ & 49.91 $_{\pm{ 3.3 }}$ & 35.22 $_{\pm{ 8.8 }}$ & 50.84 $_{\pm{ 3.8 }}$ & 39.5 $_{\pm{ 3.2 }}$\\
On-EWC  & 32.42 $_{\pm{ 1.6 }}$ & 13.73 $_{\pm{ 2.9 }}$ & 31.81 $_{\pm{ 1.4 }}$ & 14.63 $_{\pm{ 1.9 }}$ & 31.9 $_{\pm{ 1.4 }}$ & 13.79 $_{\pm{ 2.7 }}$\\
ER-OBC  & 57.54 $_{\pm{ 0.6 }}$ & 43.88 $_{\pm{ 1.2 }}$ & 58.08 $_{\pm{ 1.7 }}$ & 43.87 $_{\pm{ 2.7 }}$ & 57.76 $_{\pm{ 1.4 }}$ & 43.89 $_{\pm{ 2.1 }}$\\
CLSER   & 58.7 $_{\pm{ 1.2 }}$ & 45.74 $_{\pm{ 3.3 }}$ & 59.15 $_{\pm{ 1.2 }}$ & 46.92 $_{\pm{ 2.6 }}$ & 59.66 $_{\pm{ 1.3 }}$ & 47.39 $_{\pm{ 3.3 }}$\\
CBRS    & 57.48 $_{\pm{ 1.9 }}$ & 42.86 $_{\pm{ 1.6 }}$ & 57.39 $_{\pm{ 1.5 }}$ & 42.23 $_{\pm{ 3.0 }}$ & 58.61 $_{\pm{ 1.2 }}$ & 43.62 $_{\pm{ 3.5 }}$\\
VRMCL   & \underline{62.13} $_{\pm{ 2.9 }}$ & \underline{50.42} $_{\pm{ 2.4 }}$ & \underline{61.41} $_{\pm{ 3.8 }}$ & \underline{49.5} $_{\pm{ 3.0 }}$ & \underline{62.31} $_{\pm{ 2.6 }}$ & \underline{50.17} $_{\pm{ 2.3 }}$\\
PCL & \textbf{63.25} $_{\pm{ 0.8 }}$ & \textbf{52.73} $_{\pm{ 1.2 }}$ & \textbf{62.96} $_{\pm{ 2.1 }}$ & \textbf{51.67} $_{\pm{ 2.9 }}$ & \textbf{63.48} $_{\pm{ 1.0 }}$ & \textbf{52.09} $_{\pm{ 1.5 }}$\\
\bottomrule
    \end{tabular}
    \label{tab:seqcifar10imbalance}
\end{table*}

\begin{table*}[ht]
    \centering
    \caption{Sequential CIFAR-100 trained with small buffer budget $|\mathcal{M}| \in \{50,100,150\}$. Methods with the best average performance are \textbf{bolded}, and the second best are \underline{underlined}.}
    \begin{tabular}{@{}llllllll@{}}
    \toprule
        & \multicolumn{2}{c}{$|\mathcal{M}=50|$} & \multicolumn{2}{c}{$|\mathcal{M}=100|$} & \multicolumn{2}{c}{$|\mathcal{M}=150|$} \\ \toprule
        & AAA & Acc & AAA & Acc & AAA & Acc \\ \midrule
SGD     & 7.91 $_{\pm{ 0.3 }}$ & 3.99 $_{\pm{ 0.5 }}$ & 7.91 $_{\pm{ 0.3 }}$ & 3.99 $_{\pm{ 0.5 }}$ & 7.91 $_{\pm{ 0.3 }}$ & 3.99 $_{\pm{ 0.5 }}$\\
SI      & 7.99 $_{\pm{ 0.6 }}$ & 3.54 $_{\pm{ 0.7 }}$ & 7.99 $_{\pm{ 0.6 }}$ & 3.54 $_{\pm{ 0.7 }}$ & 7.99 $_{\pm{ 0.6 }}$ & 3.54 $_{\pm{ 0.7 }}$\\
On-EWC  & 9.59 $_{\pm{ 1.0 }}$ & 3.98 $_{\pm{ 0.9 }}$ & 9.59 $_{\pm{ 1.0 }}$ & 3.98 $_{\pm{ 0.9 }}$ & 9.59 $_{\pm{ 1.0 }}$ & 3.98 $_{\pm{ 0.9 }}$\\
LWF     & 8.12 $_{\pm{ 0.6 }}$ & 4.69 $_{\pm{ 0.7 }}$ & 8.12 $_{\pm{ 0.6 }}$ & 4.69 $_{\pm{ 0.7 }}$ & 8.12 $_{\pm{ 0.6 }}$ & 4.69 $_{\pm{ 0.7 }}$\\
ER-OBC  & 14.3 $_{\pm{ 0.6 }}$ & 6.49 $_{\pm{ 0.3 }}$ & 15.7 $_{\pm{ 0.5 }}$ & 7.45 $_{\pm{ 0.3 }}$ & 16.65 $_{\pm{ 0.7 }}$ & 7.83 $_{\pm{ 0.3 }}$\\
CLSER   & \underline{16.2} $_{\pm{ 0.4 }}$ & \underline{7.59} $_{\pm{ 0.3 }}$ & \underline{17.5} $_{\pm{ 0.9 }}$ & \underline{8.92} $_{\pm{ 0.4 }}$ & \underline{18.86} $_{\pm{ 0.7 }}$ & \underline{9.46} $_{\pm{ 0.4 }}$\\
VRMCL   & 9.11 $_{\pm{ 1.2 }}$ & 3.36 $_{\pm{ 0.8 }}$ & 9.8 $_{\pm{ 0.2 }}$ & 3.95 $_{\pm{ 0.1 }}$ & 11.66 $_{\pm{ 0.6 }}$ & 5.18 $_{\pm{ 0.4 }}$\\
PCL & \textbf{18.71} $_{\pm{ 1.4 }}$ & \textbf{9.36} $_{\pm{ 1.7 }}$ & \textbf{19.14} $_{\pm{ 0.5 }}$ & \textbf{9.71} $_{\pm{ 0.8 }}$ & \textbf{20.0} $_{\pm{ 0.6 }}$ & \textbf{10.47} $_{\pm{ 1.2 }}$\\
\bottomrule
    \end{tabular}
    \label{tab:seqcifar100diffbuff_small}
\end{table*}

\section{EVALUATION}
In this section, we empirically evaluate the performance of the proposed Parabolic Continual Learners (PCL) on benchmark datasets. 
We evaluate the performance of PCL in continual classification settings and test its behavior over a wide range of buffer sizes. 
We also test the robustness of the proposed method under label noise and its generalization ability in an imbalance classification scenario. 
We apply the Mammoth framework~\citep{buzzega2020dark} to compare different CL methods under the same datasets and experimental pipeline. 

We focus on the online, class-incremental continual learning setting where we only train one epoch for each task, and new classes are added with each subsequent task. The continual learner must learn to distinguish amongst the classes within the current task and across previous tasks. We use three datasets, Sequential CIFAR-10, Sequential CIFAR-100, and Sequential TinyImageNet. 
For Seq-CIFAR-10, we construct 5 different tasks, and each task
contains 2 classes. 
For Seq-CIFAR100, 10 tasks are derived, each composed of 10 classes. Seq-TinyImageNet is divided into 20 tasks, each containing 10 classes. 

\paragraph{Benchmarks}
Throughout the paper, we will use two main metrics to evaluate the efficacy of different methods. 
We use final averaged accuracy (Acc) across all classes once the sequential training on all tasks as the main metric to evaluate the performance. 
In the online setting, we also consider Averaged Anytime Accuracy (AAA) to evaluate the model after each task is trained. 
This is defined by the following: $\mathrm{AAA} = \frac{1}{N}\sum_j^N \mathrm{AA}_j$  where $\mathrm{AA}_j$ is the average accuracy over all classes after training of task $\tau_j$ for $N$ total tasks.

We benchmark against regularization based methods such as EWC~\citep{huszar2018note}, LWF~\citep{li2017learning}, and SI \citep{zenke2017continual}, replay based methods including GEM \citep{lopez2017gradient}, AGEM \citep{chaudhry2018efficient}, ER \citep{rolnick2019experience}, DER, DER++\citep{buzzega2020dark}, ER-OBC \citep{chrysakis2023online}, CLSER \citep{arani2022learning} and MixupER \citep{lim2024mixer} and a meta learning based methods, current SOTA in online class-incremental continual learning, VRMCL \citep{wu2024metacont}. All experiments in this paper are conducted with the ResNet18 Architecture (no pretrain) unless stated otherwise \citep{he2016deep}. All experiments are repeated over 5 trials\footnote{Averages and standard deviations are computed over 5 trials.}. We provide additional experiments including ablation studies on sampling strategies, computational efficiency comparison with VRMCL, and empirical validity of the Proposition \ref{prop:forget} and Corollary \ref{corr:general} in Appendix \ref{appendix:add_exp}. The related hyperparameters for training PCL is provided in Appendix \ref{appendix:hyperparameters}. The code to replicate these benchmarking experiments is provided in \hyperlink{https://github.com/imkeithyang/Parabolic_CL}{https://github.com/imkeithyang/Parabolic\_CL}

\subsection{Online Class Incremental Learning}
\label{sec:clresult}
In this section, we compare PCL against other continual learning methods on the task of online, class incremental learning. We evaluate the performance of these methods over 3 different datasets with buffer size $\mathcal{M} = 1000$ (see Table~\ref{tab:seqallfixbuff}). PCL outperforms all benchmarks in most of the datasets in Acc, while remaining competitive against current SOTA in AAA. In appendix~\ref{appendix:runtime}, we show that PCL achieves this competitive result against the state of the art with much-improved training efficiency. With Seq-CIFAR100, we also tested different methods with different buffer sizes $|\mathcal{M}|\in \{200, 600, 1000\}$ (see Table~\ref{tab:seqcifar100diffbuff}). PCL remains comparative to VRMCL in both metrics across three different buffer size. 

\subsection{Label Corrupted Continual Learning}
In the label corruption setting, we randomly corrupt 50\% of the training label and train all methods similarly as in Section \ref{sec:clresult} with three different datasets and Seq-CIFAR100 tested with different buffer sizes. Note that if data with a corrupted label are sampled into the buffer, the corrupted label is also stored in the buffer. We show that PCL is highly robust, since PCL is one of the methods least affected by corrupted labels and remains highly competitive across different datasets and different buffer sizes. 

\subsection{Imbalanced Continual Learning}
Finally, we examine the generalization ability of PCL by training with imbalanced data under the continual learning scenario. We follow~\citet{wu2024metacont} and benchmark PCL with imbalance sequential CIFAR-10 data with imbalance factor $\gamma = 2$ under 3 different settings. In the \textit{Normal} setting, the total number of samples in tasks decreases sequentially; the \textit{Reverse} setting is the opposite; and the total number of samples of each task in the \textit{Random} setting does not follow any sequential pattern. For the imbalance setting, we include an additional benchmark CBRS~\citep{chrysakis2020online} that specializes in online imbalance learning. In table \ref{tab:seqcifar10imbalance}, we show that PCL outperforms all benchmarked algorithms. This suggests that PCL is able to construct a meaningful domain containing all the classes, leading to a well optimized learner under the imbalance scenario.

\subsection{Experience replay with small buffer size}
We also evaluate a selected set of CL methods that have competitive performance and robustness under decreased buffer sizes of $|\mathcal{M}|\in \{50,100,150\}$. We also compare these CL methods with those that do not need a buffer during training. As demonstrated in Table \ref{tab:seqcifar100diffbuff_small}, on the Seq CIFAR-100 dataset and a small buffer size, PCL greatly outperformed all other methods in all metrics, suggesting that PCL is highly effective when only a small memory budget is available. Combining this result with PCL's robustness under 50\% corrupted training label, and the state-of-the-art performance in the imbalance data regime, we have empirically shown that PCL is a highly robust continual learning algorithm that can be highly applicable in real world environment. 

\section{DISCUSSION}
In this work we described a continual learning algorithm based on constraining the expected loss over new realizations of data to be the solution of a parabolic PDE.
We do this using the stochastic representation of parabolic PDEs to minimize a loss over the expected continual learning loss. 
The empirical results suggest the method is competitive with existing methods and outperforms baselines on tasks with corrupted data. 
Moreover, we introduce a new perspective and framework for analyzing continual loss profiles, allowing for further theoretical analysis. 

\paragraph{Limitations}
There are a number of limitations regarding this framework.
First, the bound on the growth of the solution is only an upper bound for a fixed $\ell_{f_\theta}$. 
In practice, it may be helpful to understand how the loss changes under the new optimization procedure. Sampling Brownian bridges incurs extra computational complexity compared to other experience replay based algorithms. 
Moreover, connections between the PDE and the Hamilton-Jacobi-Bellman equation should exist and could possibly be used to provide further insights on the optimization procedure. Also, PCL's empirical setting mainly focuses on continual vision tasks. However, with the rise of complex multimodal application, PCL should be further developed to consider both vision and language in a continual learning environment \citep{srinivasan2022climb}.
We leave these as future works. 

\acknowledgments

Haoming Yang and Vahid Tarokh were supported in part by the Air Force Office of Scientific Research under award number FA9550-22-1-0315.

\newpage

\bibliographystyle{plainnat}
\bibliography{ref}

\clearpage
\newpage

\onecolumn
\appendix

\section*{Supplementary Material}

\section{Additional Experiments}
\label{appendix:add_exp}
In this section we highlight a few different comparisons including computation efficiency comparison between PCL and VRMCL and ablation studies testing different sampling strategies to obtain boundary data in the buffer. We also test the validity of the proposed bounds in Proposition \ref{prop:forget} and Corollary \ref{corr:general}. 

\subsection{Ablation Study}
In this ablation study, we experiments with different Brownian Bridge (BB) sampling strategies as well as Buffer sampling strategy. Specifically for BB sampling strategies, we compare a single Brownian bridge (one BB); and tempering variance strategy (BB Tempering) where the variance of the Brownian bridge is not a constant, but a linear interpolation between the gradient norm of loss wrt data (i.e $||\nabla_x\ell(f_\theta(x) y)||_2$); sampling endpoints based on inversely sorted euclidean distance between the batch of data (Eu endpoints). For the buffer sampling strategy, reservoir sampling is still employed, but we filter the data such that either the data with the highest loss value (Max Loss), the lowest loss value (Min Loss), or the medium loss value (Middle 50\% Loss) is sampled into the buffer. Table \ref{tab:ablaseqallfixbuff} and \ref{tab:ablaseqcifar100diffbuff} showcase the AAA and Acc of different sampling strategies for different datasets and different buffer sizes. We show that there is no significant difference between each of the sampling strategies, and thus suggests that the PDE constrained optimization is robust and flexible to different conditions. 

\begin{table*}[htbp]
    \centering
    \caption{Sequential CIFAR10, sequential CIFAR100 and sequential TinyImageNet at buffer size $\mathcal{M} = 1000$. Methods with the best average performance are \textbf{bolded}.} 
    \begin{tabular}{@{}llllllll@{}}
    \toprule
          & \multicolumn{2}{c}{Seq-CIFAR10} & \multicolumn{2}{c}{Seq-CIFAR100} & \multicolumn{2}{c}{Seq-TinyImageNet}    \\ \midrule
          & AAA & Acc & AAA & Acc & AAA & Acc \\ \midrule
One BB    & 63.74 $_{\pm{ 2.0 }}$ & 53.23 $_{\pm{ 3.5 }}$ & 24.61 $_{\pm{ 1.3 }}$ & 20.25 $_{\pm{ 0.8 }}$ & 17.02 $_{\pm{ 0.5 }}$ & 9.52 $_{\pm{ 0.8 }}$\\
BB Tempering  & 63.64 $_{\pm{ 1.9 }}$ & 54.05 $_{\pm{ 3.9 }}$ & 26.01 $_{\pm{ 0.5 }}$ & 21.12 $_{\pm{ 1.9 }}$ & 17.69 $_{\pm{ 0.8 }}$ & 10.43 $_{\pm{ 0.6 }}$\\
Eu endpoints  & 63.99 $_{\pm{ 2.3 }}$ & 52.8 $_{\pm{ 4.0 }}$ & 24.68 $_{\pm{ 1.1 }}$ & 19.54 $_{\pm{ 1.2 }}$ & 16.79 $_{\pm{ 0.8 }}$ & 9.8 $_{\pm{ 0.8 }}$\\
Max Loss      & 64.45 $_{\pm{ 2.0 }}$ & 55.85 $_{\pm{ 2.3 }}$ & 25.87 $_{\pm{ 1.1 }}$ & 21.24 $_{\pm{ 0.5 }}$ & 16.65 $_{\pm{ 0.5 }}$ & 10.15 $_{\pm{ 1.4 }}$\\
Min Loss      & 64.78 $_{\pm{ 1.9 }}$ & 56.27 $_{\pm{ 1.5 }}$ & 25.2 $_{\pm{ 1.2 }}$ & 20.75 $_{\pm{ 0.6 }}$ & 17.91 $_{\pm{ 0.0 }}$ & 9.41 $_{\pm{ 0.0 }}$\\
Middle 50\% Loss  & 63.18 $_{\pm{ 2.5 }}$ & 53.34 $_{\pm{ 4.6 }}$ & 24.97 $_{\pm{ 1.2 }}$ & 18.54 $_{\pm{ 2.8 }}$ & 16.61 $_{\pm{ 0.7 }}$ & 9.93 $_{\pm{ 1.1 }}$\\
PCL             & 64.55 $_{\pm{ 1.9 }}$ & 54.98 $_{\pm{ 4.1 }}$ & 25.82 $_{\pm{ 1.2 }}$ & 20.54 $_{\pm{ 0.9 }}$ & 16.99 $_{\pm{ 0.5 }}$ & 10.12 $_{\pm{ 1.2 }}$\\
\bottomrule
    \end{tabular}
    \label{tab:ablaseqallfixbuff}
\end{table*}

\begin{table*}[htbp]
    \centering
    \caption{Sequential CIFAR-100 trained with buffer size $\mathcal{M} \in \{200,600,1000\}$. Methods with the best average performance are \textbf{bolded}.}
    \begin{tabular}{@{}llllllll@{}}
    \toprule
            & \multicolumn{2}{c}{$\mathcal{M}=200$} & \multicolumn{2}{c}{$\mathcal{M}=600$} & \multicolumn{2}{c}{$\mathcal{M}=1000$} \\ \midrule
            & AAA & Acc & AAA & Acc & AAA & Acc \\ \midrule
One BB      & 20.13 $_{\pm{ 0.5 }}$ & 12.47 $_{\pm{ 1.1 }}$ & 24.34 $_{\pm{ 1.8 }}$ & 17.89 $_{\pm{ 1.0 }}$ & 24.61 $_{\pm{ 1.3 }}$ & 20.25 $_{\pm{ 0.8 }}$ \\
BB Tempering   & 21.04 $_{\pm{ 0.8 }}$ & 13.04 $_{\pm{ 0.6 }}$ & 25.06 $_{\pm{ 1.3 }}$ & 19.1 $_{\pm{ 1.7 }}$ & 26.01 $_{\pm{ 0.5 }}$ & 21.12 $_{\pm{ 1.9 }}$\\
Eu endpoints   & 20.63 $_{\pm{ 0.9 }}$ & 11.74 $_{\pm{ 1.1 }}$ & 23.82 $_{\pm{ 1.7 }}$ & 17.15 $_{\pm{ 2.7 }}$ & 24.68 $_{\pm{ 1.1 }}$ & 19.54 $_{\pm{ 1.2 }}$\\
Max Loss       & 21.14 $_{\pm{ 0.7 }}$ & 12.2 $_{\pm{ 1.1 }}$ & 24.09 $_{\pm{ 1.2 }}$ & 18.29 $_{\pm{ 1.1 }}$ & 25.87 $_{\pm{ 1.1 }}$ & 21.24 $_{\pm{ 0.5 }}$\\
Min Loss       & 20.72 $_{\pm{ 1.3 }}$ & 11.86 $_{\pm{ 0.6 }}$ & 24.07 $_{\pm{ 0.9 }}$ & 16.77 $_{\pm{ 1.3 }}$ & 25.2 $_{\pm{ 1.2 }}$ & 20.75 $_{\pm{ 0.6 }}$\\
Middle 50\% Loss  & 21.06 $_{\pm{ 0.7 }}$ & 12.27 $_{\pm{ 1.3 }}$ & 23.85 $_{\pm{ 1.1 }}$ & 17.08 $_{\pm{ 2.2 }}$ & 24.97 $_{\pm{ 1.2 }}$ & 18.54 $_{\pm{ 2.8 }}$\\
PCL & 20.51 $_{\pm{ 0.7 }}$ & 12.52 $_{\pm{ 0.7 }}$ & 24.62 $_{\pm{ 1.5 }}$ & \textbf{17.63} $_{\pm{ 1.6 }}$ & 25.82 $_{\pm{ 1.2 }}$ & \textbf{20.54} $_{\pm{ 0.9 }}$\\
\bottomrule
    \end{tabular}
    \label{tab:ablaseqcifar100diffbuff}
\end{table*}

\subsection{Empirical Evaluation of Bounds}
We considered the following example on CIFAR-10 to test how well the bound in Proposition 1 empirically works. For this experiment, we obtain the time series (see Table \ref{tab:bound1}) of losses for data in the buffer after the training each task and compare it to the loss of the first task after each additional task is added. Table \ref{tab:bound1} suggests that the loss of the first task is bounded by the loss on the buffer, as expected and the forgetting error is bounded.
\begin{table}[htbp]
    \centering
    \caption{Comparing loss for data in buffer and loss of the first task after each additional task in a continual learning environment.}
    \begin{tabular}{llllll}
    \toprule
         & After Training Task 1 & Task 2 & Task 3 & Task 4 & Task 5 \\ \midrule
    Avg Loss of data in $\mathcal{M}$ & 0.5077 & 0.9894 & 1.3542 & 1.4373 & 1.4208 \\
    Avg Loss of Task 1  & 0.4163 & 0.2519 & 0.2888 & 0.3983 & 0.2855\\ \bottomrule
    \end{tabular}
    \label{tab:bound1}
\end{table}

Similarly for For Corollary 1, we repeat the experiment but we compare the loss of the last task loss to the loss of the buffer. Results presented in Table \ref{tab:bound2} also suggest that the bounds are maintained for Corollary 1.
\begin{table}[htbp]
    \centering
    \caption{Comparing loss for data in buffer and loss of the last task after each additional task in a continual learning environment.}
    \begin{tabular}{llllll}
    \toprule
         & After Training Task 1 & Task 2 & Task 3 & Task 4 & Task 5 \\ \midrule
    Avg Loss of data in $\mathcal{M}$ & 0.0 & 2.704e-08 & 2.082e-05 & 1.526e-05 & 1.937e-05 \\
    Avg Loss of Task 5  &  0.7139 & 0.6991 & 0.7061 & 0.7374 & 0.3960 \\ \bottomrule
    \end{tabular}
    \label{tab:bound2}
\end{table}

\subsection{Computational Efficiency Comparison}
\label{appendix:runtime}
In this section, we compare the computational efficiency between our PCL and the current state of the online continual learning methods VRMCL \citep{wu2024metacont}. We benchmark the training runtime of both methods on Seq-CIFAR100 with a buffer size of 1000. Both methods are trained on a single NVIDIA RTX 6000 GPU. VRMCL's training time on this benchmark is 972.52 ($\pm 10.27$) seconds averaged ($\pm$ SD) over 5 trials; PCL achieved a 164.99 ($\pm 2.62$) seconds averaged($\pm$ SD) over 5 trials. PCL is more than 5 times more efficient during training to achieve comparable state-of-the-art online continual learning results. 

\section{Hyperparameters}
\label{appendix:hyperparameters}
Our training hyperparameters for the PCL for all experiments remain the same with a learning rate of 0.08, batch size of 32, buffer batch size of 32, Brownian bridge diffusion coefficient for the data is 0.03, and for the respective label is 0.01. We used a terminal time $T=1$ to sample Brownian bridges, one Brownian bridge is sampled for each pair of endpoints, and there are 5 timesteps (including the two endpoints) for each Brownian bridge. A grid search is applied to select important hyperparameters including the learning rate and Brownian bridge diffusion coefficients. 

\section{Dataset and Implementation}
\paragraph{Sequential CIFAR-10}
Sequential CIFAR-10 is derived from CIFAR-10. The CIFAR-10 dataset contains 10 classes, and they are divided into 5 different binary classification tasks. The tasks are sequentially ordered to train the model.

\paragraph{Sequential CIFAR-100}
Sequential CIFAR-100 is derived from CIFAR-100, which contains 100 classes. The dataset is then divided into 10 10-class classification task to sequentially train the classifier.

\paragraph{Sequential tiny-imagenet}
Sequential tiny-imagenet is derived from tiny-imagenet, which contains 200 classes. The dataset is divided into 20 10-class classification tasks. Similar to Seq CIFAR-100 and Seq CIFAR-10, these 20 tasks are used to sequentially train the classifier. 

\paragraph{Imbalanced Experiments}
We conducted imbalanced online continual learning experiments similar to \citep{wu2024meta} as we chose the imbalance ratio $\gamma$ = 2 and followed similar imbalance ordering. The imbalance experiment is done exclusively on Seq CIFAR-10, which contains 5 tasks and 2 classes per task. Two classes with sample sizes and orders listed below are grouped to create a task. For the normal order,the number of samples for each class across all tasks is [5000, 4629, 4286, 3968, 3674, 3401, 3149, 2916, 2700, 2500].  We also considered the reversed version, where the number of samples for each class is [2500, 2700, 2916, 3149, 3401, 3674, 3968, 4286, 4629, 5000]. For the random order, we sampled a random version with [2700, 2500, 5000, 4286, 3674, 3968, 3149, 3401, 2916, 4629] samples for the classes. 

\section{Proofs}
\label{appendix:proof}
\subsection{Forgetting Error}
The forgetting error is obtained by using a maximum principle.  

\textbf{Proposition \ref{prop:forget} }[Upper bound on expected forgetting error]\textit{
    Consider $u(x_{t-\tau},t-\tau)$ for $x_{t-\tau} \in \mathcal{C}_{\mathcal{M}_t}$ and $\tau \geq 0$.
    Additionally suppose that the learning landscape satisfies the following parabolic PDE $$\frac{\partial u}{\partial t} = -\nabla^2 u(x,t) +\ell_{f_\theta}(x) \quad on \quad (x, t) \in \mathcal{C}_{\mathcal{M}_t} \times [0, T]$$
    Then the following inequality holds: $$u(x, t-\tau) \leq \max_{x, s \in \mathcal{M}_t \cup \partial \mathcal{C}_{M_T} } u(x, s).$$
}
\begin{proof}
The result follows from the maximum principle and can be easily seen from the stochastic representation of the solution.
Since 
\begin{align*}
u(x, t-\tau) &= -\mathbb{E}\left[\int_{t-\tau}^{t} \ell_{f_\theta}(x_s)\mathrm{d} s \mid x_{t-\tau} = x \right] + \mathbb{E}\left[ \ell_{f_\theta}(x_t) \mid x_{t-\tau} = x\right] \\
&\leq \mathbb{E}\left[ \ell_{f_\theta}(x_s) \mid x_{t-\tau} = x\right] \\
&\leq \max_{x,s \in \mathcal{M}_t^{\epsilon} \cup \partial \mathcal{C}_{\mathcal{M}_t^{(\epsilon)}}} \ell_{f_\theta}(x_s) \\
&= \max_{x,s \in \mathcal{M}_t^{(\epsilon)} \cup \partial \mathcal{C}_{\mathcal{M}_t^{(\epsilon)}}} u(x,s).
\end{align*}
Since $\ell$ is assumed to be positive, the terminal condition of $\ell_{f_\theta}(x)$ is the value on the boundary and the maximal value of the function $u$. 
\end{proof}

\subsection{Generalization Error}
For the generalization error, which describes the rate at which the error accumulates, we consider the opposite approach for the forward PDE.
We restate the bound here and include the proof below.

\textbf{Corollary \ref{corr:general}}[Lower bound on expected generalization error]
    \textit{Consider $u(x_{t+\tau},t+\tau)$ for $x_{t+\tau} \in \mathcal{C}_{\mathcal{M}_t}$ and $\tau \geq 0$. Suppose that the expected continual learning loss $u$ satisfies the following parabolic PDE
    $$\frac{\partial u}{\partial t} = \nabla^2 u(x,t) 
    + \ell_{f_\theta}(x) \quad on \quad (x, t) \in \mathcal{C}_{\mathcal{M}_t} \times [T, \infty)$$
    and that the function $\ell_{f_\theta}(x)$ is $C-$Lipshitz in $x$. 
    Then the following inequality holds:
    $$\min_{x, s \in \mathcal{M}_t \cup \partial \mathcal{C}_{M_T} } u(x, s) \leq u(x, t+\tau) \leq  C\tau + \max_{ x \in \mathcal{C}_{M_T}} u(x,s) .$$
}
\begin{proof}
First note that we assume that the loss conditioned on $f_\theta$ is non-negative for all points within our domain.
We can then write the PDE as $\frac{\partial u}{\partial t} + \nabla^2 u \geq 0.$
From~\citet{pardoux2014stochastic} PDEs of this form satisfy a minimum principle on the boundary data which gives us the left hand side of the of the inequality. 
This can also be seen since the solution is given as the following expectation:
\begin{align*}
u(x, t+\tau) &= \mathbb{E}\left[\int_t^{t+\tau} \ell_{f_\theta}(x_s)\mathrm{d} s \mid x_t = x\right] + \mathbb{E}\left[ \ell_{f_\theta}(x_s) \mid x_t = x\right] \\
&\geq \mathbb{E}\left[ \ell_{f_\theta}(x_s) \mid x_t = x \right] \\
&\geq \min_{x,s \in \mathcal{M}_t^{(\epsilon)} \cup \partial \mathcal{C}_{\mathcal{M}_T^{(\epsilon)}}} \ell_{f_\theta}(x_s) \\
&= \min_{x,s \in \mathcal{M}_t^{(\epsilon)} \cup \partial \mathcal{C}_{\mathcal{M}_T^{(\epsilon)}}} u(x,s).
\end{align*}

For the upper bound, we use the stochastic representation again.
\begin{align*}
u(x, t+\tau) &= \mathbb{E}\left[\int_t^{t+\tau} \ell_{f_\theta}(x_s)\mathrm{d} s\right] + \mathbb{E}\left[ \ell_{f_\theta}(x_s)\right] \\
&\leq C\tau + \mathbb{E}\left[ \ell_{f_\theta}(x_s)\right] \quad \text{($\ell_{f_\theta}$ is $C$-Lipschitz)} \\
&\leq C\tau + \max_{x,s \in \mathcal{M}_t^{(\epsilon)} \cup \partial \mathcal{C}_{\mathcal{M}_T^{(\epsilon)}}} \ell_{f_\theta}(x_s) \\
&= C \tau + \max_{x,s \in \mathcal{M}_t^{(\epsilon)} \cup \partial \mathcal{C}_{\mathcal{M}_T^{(\epsilon)}}} u(x,s).
\end{align*}
Putting these together gives us the desired bound.
\end{proof}

\subsection{Bound Using First Order Derivatives}

\textbf{Remark 1} [An Upper Bound on the Negative Log Loss]
\textit{Suppose we optimize the continual loss profile under the following PDE
\begin{equation}
\frac{\partial u^\varphi }{\partial t} = -\nabla \varphi(x)^\top \nabla u^\varphi (x) + \frac12 \nabla^2 u^\varphi (x) + \ell_{f_\theta}(x)
\label{eq:pde_first_app}
\end{equation}
where $\varphi(x)$ is convex with minimum at $x^\star$. 
Then the negative log loss of the solution $u^\varphi $ for any given $\ell_{f_\theta}$ is bounded from above by a value greater than or equal to the upper bound of the solution to~\eqref{eq:pde} under the same $\ell_{f_\theta}$.
That is, optimizing~\eqref{eq:pde_first_app} has a greater upper bound than the PDE without the linear term for all points except $x^\star$ where the bounds are equal.
}

\begin{proof}
    We will use the stochastic representation again for this remark.
    In this case, we will use a change of measure to describe the influence of the drift $\mu$, which in this case $\mu = \nabla \varphi(x)$.
    We additionally assume that $\mu$ satisfies the following criteria:
    The first is Novikov's citeria which guarantees that the change of measure exists; 
    The second is that $\mu$ is adapted to the filtration generated by $X_t$.
    The change of measure is given by
    \begin{equation}
    \frac{\mathrm{d}P_\mu}{\mathrm{d}Q} := \exp \left (\int_0^{t\wedge \tau_{\mathcal{M}^{(\epsilon)}}}  \mu^\top(X_s) \mathrm{d} X_s - \frac12 \int_0^{t\wedge \tau_{\mathcal{M}^{(\epsilon)}}}\mu^\top \mu(X_s) \mathrm{d}s \right ).
    \label{eq:com}
\end{equation}
The solution to the PDE by its stochastic representation is given by
$$
u^\mu(x,t) = \mathbb{E} \left [ \left(\ell_{f_\theta}(X_{t \wedge \tau_{\mathcal{M}^{(\epsilon)}}} ) + \int_0^{t\wedge \tau_{\mathcal{M}^{(\epsilon)}}} \ell_{f_\theta}(X_s) \mathrm{d}s \right) \frac{\mathrm{d}P_\mu}{\mathrm{d}Q}\mid X_0 = x\right ].
$$
Applying Jensen's inequality, we get the following
$$
-\log u^\mu(x,t) \leq \mathbb{E} \left [ -\log \ell_{f_\theta}(X_{t \wedge \tau_{\mathcal{M}^{(\epsilon)}}} ) - \log \frac{\mathrm{d}P_\mu}{\mathrm{d}Q} \mid X_0 = x\right ] + \mathbb{E}\left [ - \log \int_0^{t\wedge \tau_{\mathcal{M}^{(\epsilon)}}} \ell_{f_\theta}(X_s) \mathrm{d}s  - \log \frac{\mathrm{d}P_\mu}{\mathrm{d}Q} \mid X_0 = x\right ]
$$
The solution without the added bias is given by 
$$
-\log u \leq \mathbb{E} \left [ -\log \ell_{f_\theta}(X_{t \wedge \tau_{\mathcal{M}^{(\epsilon)}}} )\mid X_0 = x\right ] - \mathbb{E}\left [ \log \int_0^{t\wedge \tau_{\mathcal{M}^{(\epsilon)}}} \ell_{f_\theta}(X_s) \mathrm{d}s  \mid X_0 = x\right ]
$$
The difference between the upper bounds is given by the log of~\eqref{eq:com} which is
$$
-2\log \frac{\mathrm{d}P_\mu}{\mathrm{d}Q} = -2\int_0^{t\wedge \tau_{\mathcal{M}^{(\epsilon)}}}  \mu^\top(X_s) \mathrm{d} X_s + \int_0^{t\wedge \tau_{\mathcal{M}^{(\epsilon)}}}\mu^\top \mu(X_s) \mathrm{d}s 
$$
Taking expectations over normal random variables, 
$$
2\mathbb{E}\left[-\log \frac{\mathrm{d}P_\mu}{\mathrm{d}Q}\right]= -2\mathbb{E}\left[\int_0^{t\wedge \tau_{\mathcal{M}^{(\epsilon)}}}  \mu^\top(X_s) \mathrm{d} X_s \right]+\mathbb{E} \left[  \int_0^{t\wedge \tau_{\mathcal{M}^{(\epsilon)}}}\mu^\top \mu(X_s) \mathrm{d}s \right]  
$$
the first term in the stochastic integral goes to zero since $X_s$ is a Brownian motion. 
This leaves us with the second squared term which is always greater than or equal to zero. 
\end{proof}

\end{document}